\documentclass[runningheads]{llncs}
\usepackage{graphicx}

\usepackage{tikz}
\usepackage{comment}
\usepackage{amsmath,amssymb} 
\usepackage{color}

\usepackage[accsupp]{axessibility}  

\usepackage{graphicx}
\usepackage{amsmath}
\usepackage{amssymb}
\usepackage{booktabs}
\usepackage{multirow}
\usepackage{caption}
\usepackage{algorithm}
\usepackage{algpseudocode}
\usepackage{tabularx}
\usepackage{subcaption}

\usepackage{xcolor}
\definecolor{citecolor}{HTML}{0071bc}
\usepackage[pagebackref,breaklinks,colorlinks,citecolor=citecolor,bookmarks=false]{hyperref}

\makeatletter
\newcommand{\multiline}[1]{%
  \begin{tabularx}{\dimexpr\linewidth-\ALG@thistlm}[t]{@{}X@{}}
    #1
  \end{tabularx}
}
\makeatother

\begin{document}
\pagestyle{headings}
\mainmatter
\def\ECCVSubNumber{6220}  

\title{Transformers as Meta-Learners for\\Implicit Neural Representations} 

\titlerunning{Transformers as Meta-Learners for Implicit Neural Representations}
%
\author{Yinbo Chen \and Xiaolong Wang}
\authorrunning{Y. Chen and X. Wang}
%
\institute{UC San Diego}
\maketitle

\begin{abstract}
Implicit Neural Representations (INRs) have emerged and shown their benefits over discrete representations in recent years. However, fitting an INR to the given observations usually requires optimization with gradient descent from scratch, which is inefficient and does not generalize well with sparse observations. To address this problem, most of the prior works train a hypernetwork that generates a single vector to modulate the INR weights, where the single vector becomes an information bottleneck that limits the reconstruction precision of the output INR. Recent work shows that the whole set of weights in INR can be precisely inferred without the single-vector bottleneck by gradient-based meta-learning. Motivated by a generalized formulation of gradient-based meta-learning, we propose a formulation that uses Transformers as hypernetworks for INRs, where it can directly build the whole set of INR weights with Transformers specialized as set-to-set mapping. We demonstrate the effectiveness of our method for building INRs in different tasks and domains, including 2D image regression and view synthesis for 3D objects. Our work draws connections between the Transformer hypernetworks and gradient-based meta-learning algorithms and we provide further analysis for understanding the generated INRs. The project page with code is at \url{https://yinboc.github.io/trans-inr/}.
\end{abstract}

\section{Introduction}

In recent years, Implicit Neural Representations (INRs) have been proposed as continuous data representations for various tasks in computer vision. With INR, data is represented as a neural function that maps continuous coordinates to signals. For example, an image can be represented as a neural function that maps 2D coordinates to RGB values, a 3D scene can be represented as a neural radiance field (NeRF~\cite{mildenhall2020nerf}) that maps 3D locations with view directions to densities and RGB values.
Compared to discrete data representations such as pixels, voxels, and meshes, INRs do not require resolution-dependent quadratic or cubic storage. Their representation capacity does not depend on grid resolution but instead on the capacity of a neural network, which may capture the underlying data structure and reduce the redundancy in representation, therefore providing a compact yet powerful continuous data representation.

However, learning the neural functions of resolution-free INRs from given observations usually requires optimization with gradient descent steps, which has several challenges:
(i) Optimization can be slow if every INR is learned independently from a random initialization;
(ii) The learned INR does not generalize well to unseen coordinates if the given observations are sparse and no strong prior is shared.

From the perspective of efficiently building INRs, previous works~\cite{sitzmann2019siren} proposed to learn a latent space where each INR can be decoded by a single vector with a hypernetwork~\cite{Ha2017HyperNetworks}. However, a single vector may not have enough capacity to capture the fine details of a complex real-world image or 3D object, while these works show promising results in generative tasks~\cite{skorokhodov2021adversarial,anokhin2021image,chan2021pi}, they do not have high precision in reconstruction tasks~\cite{chan2021pi}. The single-vector modulated INRs are mostly used for representing local tiles ~\cite{mehta2021modulated} for reconstruction. Recent works~\cite{chen2021learning,saito2019pifu,yu2021pixelnerf} revisit the grid-based discrete representations and define INRs over deep feature maps, where the capacity and storage will be resolution-dependent and the decoding is bounded by feature maps as the INRs rely on local features. Going beyond the limitation of the resolution, our work is inspired by recent works~\cite{sitzmann2019metasdf,tancik2020meta} which explore a promising direction in the intersection between gradient-based meta-learning and INRs. Without grid-based representation, these works can efficiently and precisely infer the whole set of INR weights without the single-vector bottleneck. However, the computation of higher-order derivatives and a learned fixed initialization make these methods less flexible, and gradient descent that involves sequential forward and backward passing is still necessary for obtaining INRs from given observations in these works. 

\begin{figure}[t]
    \centering
    \includegraphics[width=\linewidth]{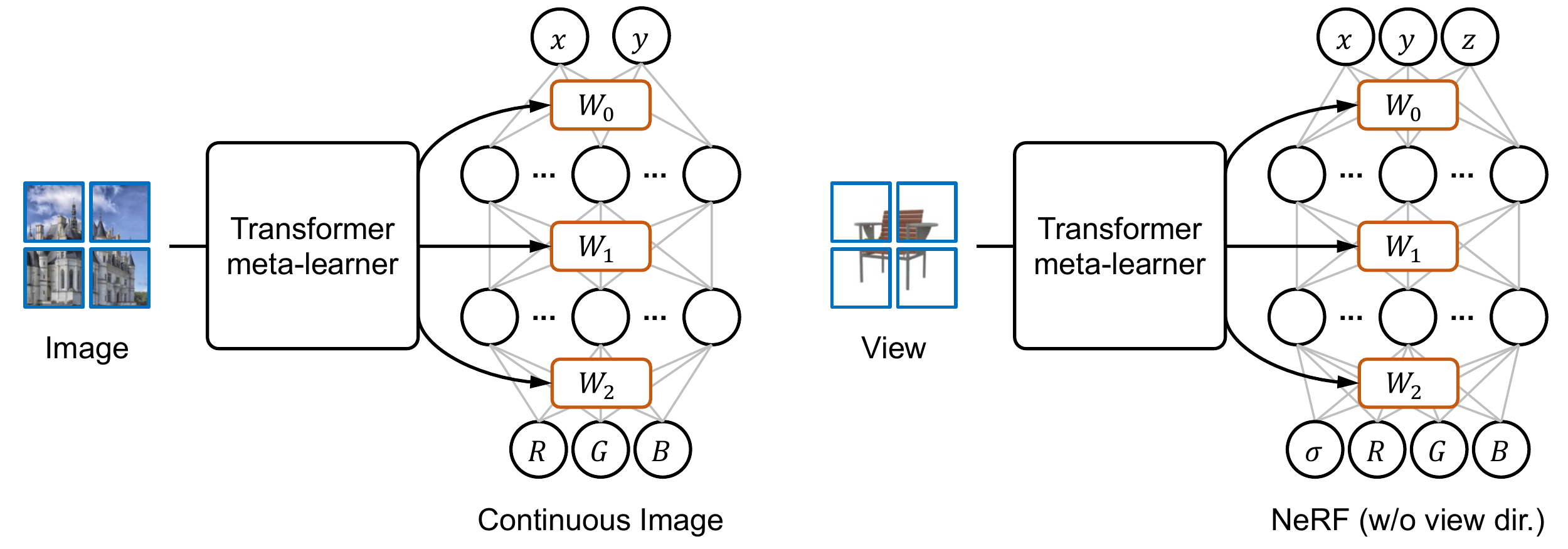}
    \caption{Implicit Neural Representation (INR) is a function that maps coordinates to signals. We propose to use Transformers as meta-learners for directly building the whole weights in INRs from given observations. Our method supports various types of INRs, such as continuous images and neural radiance fields.}
    \label{fig:teaser}
\end{figure}

Motivated by a generalized formulation of the gradient-based meta-learning methods, we propose the formulation that uses Transformers~\cite{vaswani2017attention} as effective hypernetworks for INRs (Figure~\ref{fig:teaser}). Our general idea is to use Transformers to transfer the knowledge from image observations to INR weights. Specifically, we first convert the input observations to data tokens, then we view the weights in INR as the set of column vectors in weight matrices of different layers, for which we create initialization tokens each representing one column vector. These initialization tokens are passed together with data tokens into a Transformer, and the output tokens are mapped to their corresponding location (according to the location of initialization tokens) as the weights in INRs.

We verify the effectiveness of our method for building INRs in both 2D and 3D domains, including image regression and view synthesis. We show that our approach can efficiently build INRs and outperform previous gradient-based meta-learning algorithms on reconstruction and synthesis tasks. Our further analysis shows qualitatively that the INRs built by the Transformer meta-learner may potentially exploit the data structures without explicit supervision.

To summarize, our contributions include:
\begin{itemize}
    \item We propose a Transformer hypernetwork to infer the whole weights in an INR, which removes the single-vector bottleneck and does not rely on grid-based representation or gradient computation.
    \item We draw connections between the Transformer hypernetwork and the gradient-based meta-learning for INRs.
    \item Our analysis sheds light on the structures of the generated INRs.
\end{itemize}

\section{Related Work}

\textbf{Implicit neural representation.} Implicit neural representations (INRs) have been demonstrated as flexible and compact continuous data representations in recent works. A main branch of these works use INRs for representing 3D objects or scenes, their wide applications include 3D reconstruction~\cite{deng2020nasa,genova2019learning,genova2020local,michalkiewicz2019implicit} and generation~\cite{schwarz2020graf,chan2021pi,devries2021unconstrained}. Typical examples of resolution-free INRs include DeepSDF~\cite{Park_2019_CVPR} which represents 3D shapes as a field of signed distance function, Occupancy Networks~\cite{Mescheder_2019_CVPR} and IM-NET~\cite{Chen_2019_CVPR} which represents 3D shapes as binary classification neural network that classifies each 3D coordinate for being inside or outside the shape. NeRF and its follow-up works~\cite{mildenhall2020nerf,martin2021nerf,park2020deformable,liu2020neural} are proposed to represent a scene as a neural radiance field that maps each position to a density and a view-dependent RGB value, with differentiable volumetric rendering that allows optimizing the representation from 2D views. The idea of INR has also been adapted for representing 2D images in recent works~\cite{chen2021learning,skorokhodov2021adversarial,anokhin2021image,karras2021alias}, which allows decoding for arbitrary output resolution. Several recent works observe that coordinate-based MLPs with ReLU activation may lack the capacity for representing fine details, solutions proposed to address this issue include replacing ReLU with sine activation function~\cite{sitzmann2019siren}, and using Fourier features of input coordinates~\cite{tancik2020fourfeat}.

\textbf{Hypernetworks for INRs.} A hypernetwork~\cite{Ha2017HyperNetworks} $g$ generates the weights $\theta$ for another network $f_\theta$ from some input $z$, i.e. $\theta = g(z)$. Directly building an INR from given observations will usually require performing gradient descent steps, which is inefficient and does not generalize well with sparse observations. A common way to tackle this short-come is learning a latent space for INRs~\cite{Park_2019_CVPR,Mescheder_2019_CVPR,sitzmann2019srns,sitzmann2019siren}, where each INR corresponds to a latent vector that can be decoded by a hypernetwork. Since a single vector may have limited capacity for representing the fine details (e.g. lack of details in reconstructing face image~\cite{sitzmann2019siren,chan2021pi}), many recent works~\cite{chen2021learning,genova2020local,chibane2020implicit,jiang2020local,peng2020convolutional,chabra2020deep,mehta2021modulated} address this issue by revisiting discrete representation and defining INRs with feature maps in a hybrid way, where the data still corresponds to a grid-based representation. Different from these hybrid methods, our goal is to obtain a hypernetwork that allows for building a resolution-free neural function (i.e. a global function instead of a grid-based representation).

\textbf{Meta-learning.} Learning to build a neural function from given observations is related to the topic of meta-learning, where a differentiable meta-learner is trained for inferring the weights in a neural network. Most previous works on meta-learning have been focus on few-shot learning~\cite{vinyals2016matching,NIPS2017_cb8da676,Sachin2017,sung2018learning,mishra2018a} and reinforcement learning~\cite{finn2017model,fernando2018meta,jaderberg2019human}, where a meta-learner allows fast adaptation for new observations and better generalization with few samples. Gradient-based methods is a popular branch in meta-learning algorithms, typical examples include MAML~\cite{finn2017model}, Reptile~\cite{nichol2018first}, and their extentions~\cite{antoniou2018how,fallah2020convergence,Rajeswaran2019MetaLearningWI}. A recent paper provides a comprehensive survey on meta-learning algorithms~\cite{hospedales2020meta}.

While most previous works in meta-learning aim at building a neural function for processing the data, the recent rising topic of implicit neural representation connects neural functions and data representations, which extends the idea of meta-learning with new possibilities for building neural functions that represent the data. MetaSDF~\cite{sitzmann2019metasdf} adopts a gradient-based meta-learning algorithm for learning signed distance functions, which leads to much faster convergence than standard learning. Learned Init~\cite{tancik2020meta} generalizes this idea to wider classes of INRs and shows the effectiveness of using the meta-learned initialization as encoded prior. While these works have shown promising results, their methods only learn a fixed initialization and require test-time optimization. We show that it is possible to directly build the whole INR with a Transformer meta-learner and it is more flexible than a fixed initialization.

\textbf{Transformers.} Transformers~\cite{vaswani2017attention} were initially proposed for machine translation, and has later been a state-of-the-art architecture used in various methods~\cite{devlin-etal-2019-bert,Radford2018ImprovingLU,Radford2019LanguageMA,brown2020language} in natural language processing. Recent works~\cite{dosovitskiy2021an,touvron2021training,liu2021Swin} also demonstrate the potential of Transformers for encoding visual data. In this work, we show promising results of using Transformers in meta-learning for directly inferring the whole weights in a neural function of INR.

\section{Method}

\subsection{Problem Formulation}
\label{subsec:pf}

We are interested in the problem of recovering a signal $I$ from observations $O$. The signal is a function $I: X \rightarrow Y$ defined in a bounded domain that $X\subseteq \mathbb{R}^c, Y\subseteq \mathbb{R}^d$. For instance, an image can be represented as a function that maps 2D coordinates to 3D tuples of RGB values. A 3D object or scene can be represented as a neural radiance field (NeRF)~\cite{mildenhall2020nerf}, which maps 3D locations with view directions $v$ (normalized 3D vectors) to 4D tuples that describe the densities and RGB values.

In implicit neural representation, the signal $I$ is estimated and parameterized by a neural function $f_\theta$ with $\theta$ as its weights (learnable parameters). A typical example of $f_\theta$ is a multilayer perceptron (MLP). We consider a more general class of $f_\theta$ where its weights consist of a set of matrices
\begin{equation*}
    \theta = \{W_i \mid W_i \in \mathbb{R}^{\textrm{in}_i\times \textrm{out}_i}\}_{i=0}^{m-1},
\end{equation*}
the biases to add (if exist) are merged into these matrices. Given the observations $O$, our goal is now to obtain $\theta$ that fits the signal $I$ with the neural function $f_\theta$.

Observations $O$ is a set $O=\{T_i(I)\}_{i=0}^{|O|-1}$ with transform functions $T_i$. For example, to estimate a continuous image, each pixel $i$ in the given image can be approximately viewed as $T_i(I) = I(x_i)$, where $x_i$ is the center coordinate of pixel $i$ and $I(x_i)$ are the RGB values. To estimate a 3D object with NeRF, an input view provides each pixel $i$ with its corresponding rendering ray $r_i$, that can be represented as $T_i(I) = R(I, r_i)$, where $R$ is the function renders the RGB values from ray $r_i$ in the radience field $I$.

Given the observation set $O$, estimating $I$ with the INR $f_\theta$ can be addressed by minimizing the L2 loss
\begin{equation}
    \label{eq:l_theta_o}
    L(\theta; O) = \frac{1}{|O|} \sum_{T_i\in O} \|T_i(f_\theta) - T_i(I)\|_2^2.
\end{equation}
If we assume $T_i(f_\theta)$ is differentiable to $\theta$, minimizing this loss with gradient descent steps is referred to as fitting an INR to given observations or learning the INR. The goal of a meta-learner is to efficiently find $\theta$ with given $O$ and improve the generalization of the neural function $f_\theta$.

\subsection{Motivating from gradient-based meta-learning}
\label{sec:conn_gradient}

In meta-learning, the goal is to train a meta-learner that infers the weights $\theta$ of a target network $f_\theta$ from given observations. In MAML~\cite{finn2017model}, the learnable component is an initialization $\theta_0$, $\theta=\theta_n$ is inferred by updating $\theta_0$ for $n$ steps
\begin{equation}
    \theta_{i+1} = \theta_i + (-\nabla_\theta \mathcal{L}(\theta; O)|_{\theta=\theta_i}),
\end{equation}
where $\mathcal{L}$ is the differentiable loss function computed with observations $O$. The update formula above defines a computation graph from $\theta_0$ to $\theta_n$, if the computation graph (with higher-order derivatives) is differentiable, the gradient for optimizing $\theta_n$ can be back-propagated to $\theta_0$ for training this meta-learner.

We consider a more general class of meta-learners, where its learnable components contain: (i) A learnable initialization $\theta_0$; (ii) A total number of update steps $n$; (iii) A step-specific learnable update rule $U_{\psi_i}$ (with $\psi_i$ as its parameters) that conditions on some provided data $D_i$:
\begin{equation}
    \label{eq:residual_theta}
    \theta_{i+1} = \theta_i + U_{\psi_i}(\theta_i; D_i).
\end{equation}
The meta-learning objective is applied to the final vector $\theta_n$, which is typically fitting the seen observations or generalizing to unseen observations.

\begin{figure}[t]
    \centering
    \includegraphics[width=\linewidth]{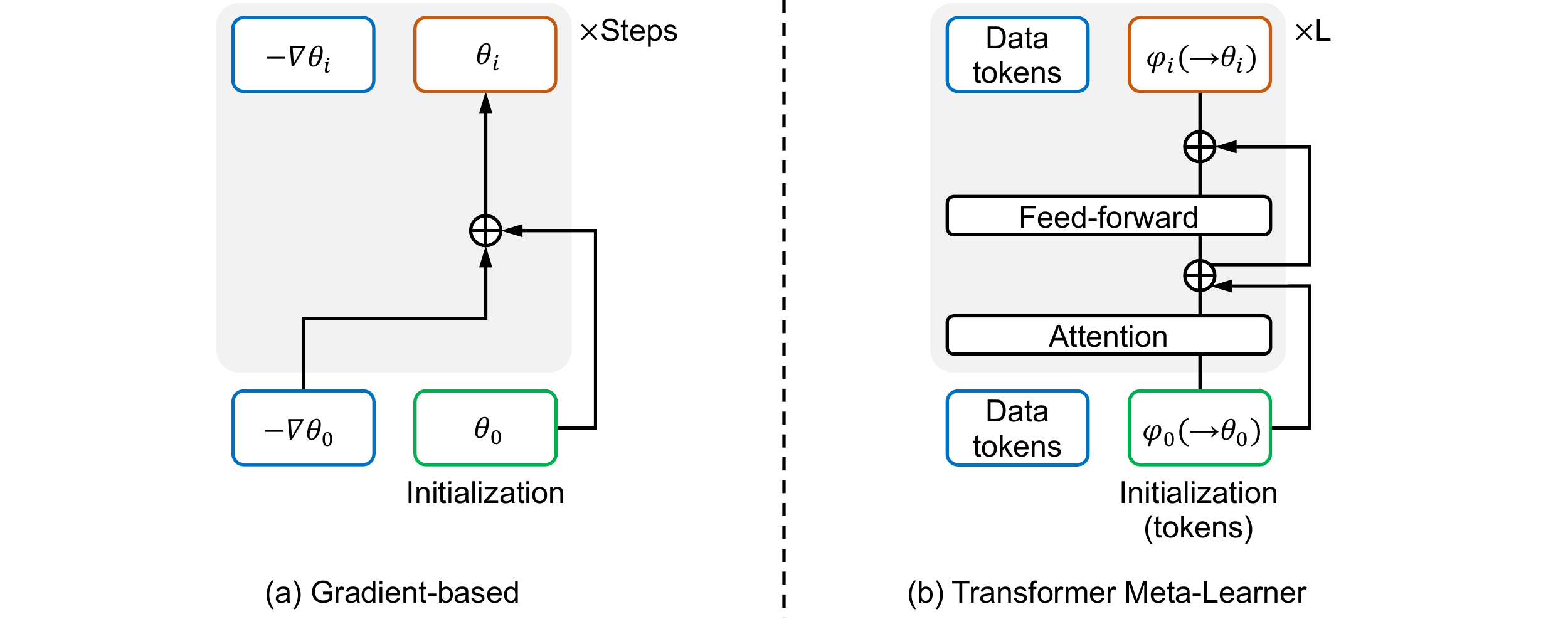}
    \caption{\textbf{Motivating from gradient-based meta-learners.} The residual link in the Transformer meta-learner shares a similar formulation as subtracting the gradients in gradient descent for updating the weights.}
    \label{fig:residual}
\end{figure}

We observe that this formulation can be naturally instantiated with a Transformer architecture. In general, we propose to represent observations as a set of data tokens, which are passed into a Transformer encoder with a set of initialization tokens that are learnable parameters defined in addition, as shown in Figure~\ref{fig:residual} (b). The computation graph with the residual link can be written as
\begin{equation}
    \label{eq:residual_varphi}
    \varphi_{i+1} = \varphi_i + U_{\psi_i}(\varphi_i; d_i),
\end{equation}
where $d_i$ are the data tokens at layer $i$, $U_{\psi_i}$ is the function that describes how the output residual is conditioned on $i$-th layer's input, i.e. the function that is composed of the attention layer and the feed-forward layer, $\varphi_0$ is the learnable initialization tokens and tokens $\varphi_i$ corresponds to the target weights $\theta_i$.

\begin{figure}[t]
    \centering
    \begin{minipage}{\linewidth}
        \includegraphics[width=\linewidth]{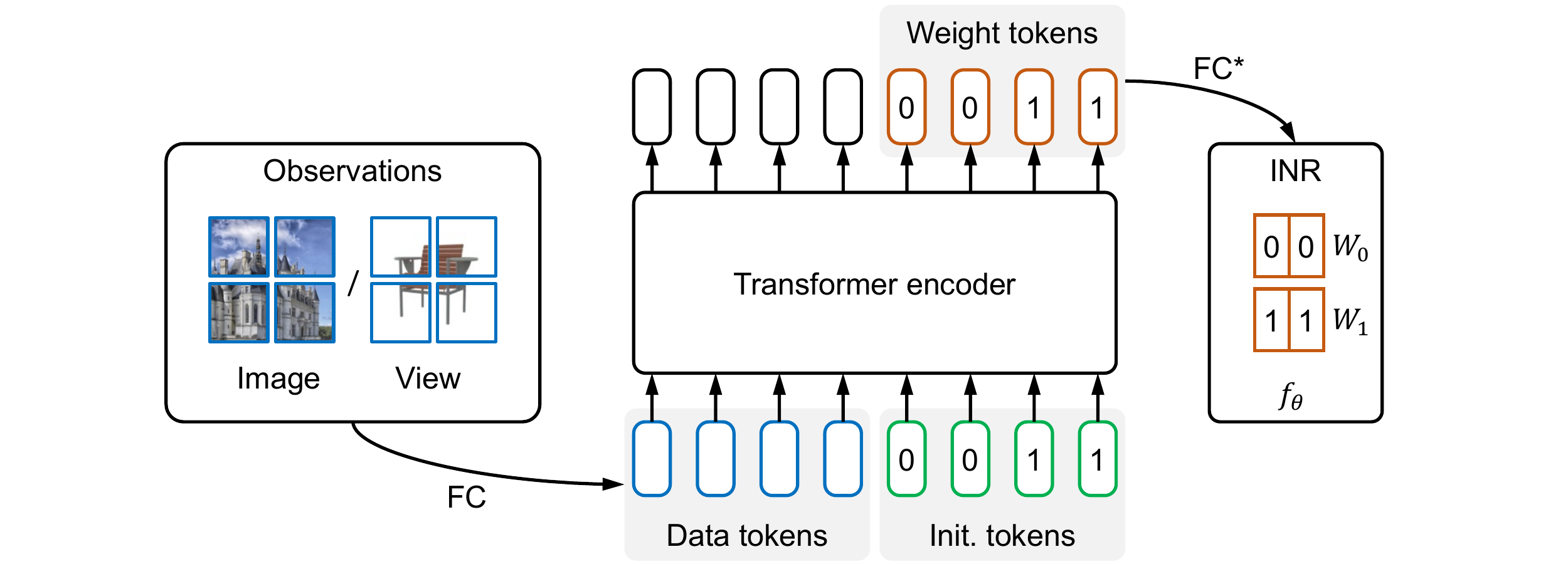}
    \end{minipage}
    \caption{\textbf{Transformers as meta-learners.} We propose to use a Transformer encoder as the meta-learner that directly builds the whole weights of an INR from given observations. The observations are split into patches and mapped to data tokens by a fully connected (FC) layer. The INR weights are viewed as the set of column vectors in weight matrices. For each column vector, we create a corresponding initialization token at the input. The data tokens and the initialization tokens are passed together into the Transformer encoder. The weight tokens are generated at the output and are mapped to the column vectors in INR weights with layerwise FCs (denoted by FC$^*$).}
    \label{fig:method}
\end{figure}

\subsection{Transformers as Meta-Learners}
\label{subsec:tm}

In this section, we introduce the details of our Transformer hypernetwork. We use Transformers to directly build the whole weights $\theta$ by transferring the knowledge from encoded information of observations $O$. Our method is demonstrated in Figure~\ref{fig:method}, in general, it represents the observations as data tokens and decodes them to weight tokens, that each weight token corresponds to some locations in the INR weights.

In practice, the observation set usually consists of images (or with given poses). We follow a similar strategy as in Vision Transformer~\cite{dosovitskiy2021an}, where the images are split into patches. The patches are flattened and then mapped by a fully connected (FC) layer to vectors in the dimension of the input to the Transformer. We denote these vectors as data tokens, i.e. the tokens that represent the observation data, which are the blue input tokens in Figure~\ref{fig:method}.

To decode for the whole INR weights $\theta = \{W_i\}_{i=0}^{m-1}$, we view each weight matrix $W_i$ as a set of column vectors, and $\theta$ can be represented as the joint of the column vector sets. For each of these column vectors, we create an initialization token (which is a learnable vector parameter) correspondingly at the input for the Transformer. In Figure~\ref{fig:method}, they are illustrated as green tokens.

These initialization tokens and data tokens are passed together into the Transformer encoder, which jointly models: (i) building features of the observations through interactions in data tokens; (ii) transferring the knowledge of observations to the weights through interactions between data tokens and initialization tokens; (iii) the relation of different weights in INR through interactions in initialization tokens.

Finally, the output vectors at the positions of the input initialization tokens are denoted as weights tokens, which are shown in Figure~\ref{fig:method} as the tokens in orange color. To map them into the INR weights, since the dimensions of the column vectors in $W_i$ can be different for different $i$, we have $m$ independent FC layers for each $i\in \{0,\dots, m-1\}$ that maps the weight tokens to their corresponding column vectors in $W_i$, which gets whole INR weights $\theta = \{W_i\}_{i=0}^{m-1}$.

To train this Transformer meta-learner, a loss is computed with regard to the INR weights $\theta$. Let $O$ denotes the observations from which we generate $\theta$, for the optimization goal of the meta-learner, the loss can be defined as $L(\theta; O)$ in Equation~\ref{eq:l_theta_o}. In tasks that require improving the generalization of the INR $f_\theta$ (e.g. view synthesis from a single input image), we sample $O'\neq O$ from the training set and compute the loss $L(\theta; O')$ instead. $L(\theta; O')$ requires the estimated $f_\theta$ to generalize to unseen observations, which explicitly adds generalization of INR as an objective.

\begin{figure}[t]
    \centering
    \includegraphics[width=\linewidth]{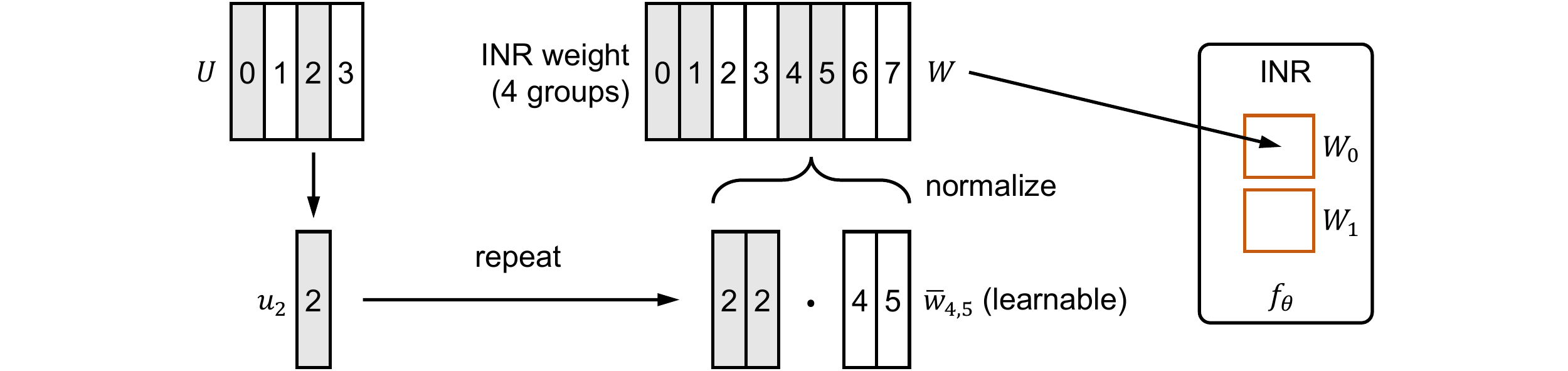}
    \caption{\textbf{Weight Grouping.} Columns in weight matrix $W$ are divided into groups, each group can be generated by a single vector. $\bar{w}_i$ are learnable vectors assigned for every column in $W$, which are independent of the input observations.}
    \label{fig:wgroup}
\end{figure}

\subsection{Weight Grouping}
\label{subsec:wg}

Assigning tokens for each column vector in weight matrices might be inefficient when the size of $\theta$ is large. To improve the efficiency and scalability of our Transformer meta-learner, we present a weight grouping strategy that offers control for the trade-off of precision and cost.

The general idea is to divide the columns in a weight matrix into groups and assign a single token for each group, as illustrated in Figure~\ref{fig:wgroup}. Specifically, let $W\in \theta$ denotes a weight matrix that can be viewed as column vectors $W=[w_0 \dots w_{r-1}]$. For weight grouping with a group size of $k$, $W$ will be defined by a new set of column vectors $U=[u_0 \dots u_{r/k -1}]$ (assume $r$ is divisible by $k$). Specifically, $w_i$ is defined by $u_{\lfloor i/k\rfloor}$ with the formula
\begin{equation*}
    w_i = \textrm{normalize}(u_{\lfloor i/k\rfloor} \cdot \bar{w_i}),
\end{equation*}
where normalize refers to L2 normalization, $\bar{w_i}$ are the learnable parameters assigned for every weight $w_i$, note that they are independent of the given observations. With this formulation, $U$ will replace $W$ as the new weights for the Transformer meta-learner to build.

The weight grouping strategy will roughly reduce the number of weight tokens by a factor of $k$, which makes it more efficient for the Transformer meta-learner to build the weights while maintaining the representation capacity of the inferred INR $f_\theta$. $\bar{w}_i$ for every column vector $w_i\in W$ are learnable vectors that do not need to be generated by the Transformer, and they make the columns vector within the same group still different from each other.

\section{Experiments}

\subsection{Image Regression}

\begin{figure}[t]
    \centering
    \begin{tabular}{cc|cc}
        $f_\theta(x)$ & GT & $f_\theta(x)$ & GT \\
        ~~\includegraphics[width=.17\linewidth]{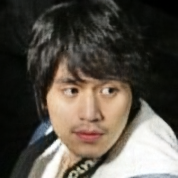}~~ & ~~\includegraphics[width=.17\linewidth]{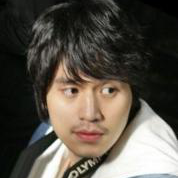}~~ &
        ~~\frame{\includegraphics[width=.17\linewidth]{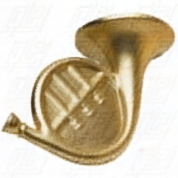}}~~ & ~~\frame{\includegraphics[width=.17\linewidth]{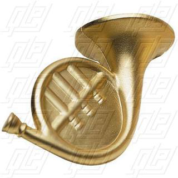}}~~ \\
        ~~\includegraphics[width=.17\linewidth]{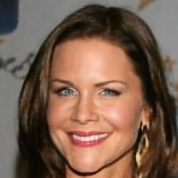}~~ & ~~\includegraphics[width=.17\linewidth]{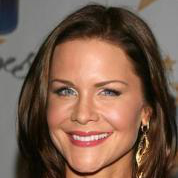}~~ &
        ~~\frame{\includegraphics[width=.17\linewidth]{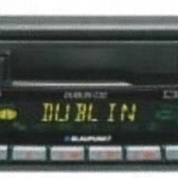}}~~ & ~~\frame{\includegraphics[width=.17\linewidth]{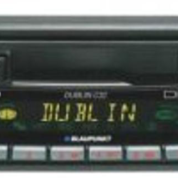}}~~ \\
    \end{tabular}
    \caption{\textbf{Qualitative results of image regression.} Our method builds the weights of $f_\theta$ that fit the observations of the target image and recovers the details of real-world images. Examples are shown on CelebA (left) and Imagenette (right), which are face images and natural images of general objects.}
    \label{fig:imgrec}
    \vspace{-1em}
\end{figure}

Image regression is a basic task commonly used for evaluating the representation capacity of INRs in recent works~\cite{sitzmann2019siren,tancik2020meta}. In image regression, a target image $J$ is sampled from an image distribution $J\sim \mathcal{J}$. An INR $f_\theta$ is a neural network that takes as input a 2D coordinate in the image and outputs the RGB value. The goal is to infer the weights $\theta$ in INR $f_\theta$ for a given target image $J$ so that $f_\theta$ can reconstruct $J$ by outputting the RGB values at the center coordinates of pixels in $J$. Unlike previous works that perform gradient descent steps to optimize the INR weights for given observations, our goal is to use a Transformer to directly generate the INR that can fit the pixel values in the target image without test-time optimization.

\textbf{Setup.} We follow the datasets of real-world images used in recent work~\cite{tancik2020meta}. \textit{CelebA}~\cite{liu2015deep} is a large-scale dataset of face images. It contains about 202K images of celebrities, which are split into 162K, 20K, and 20K images as training, validation, and test sets. \textit{Imagenette}~\cite{howard2020imagenette} is a dataset of common objects. It is a subset of 10 classes chosen from the 1K classes in ImageNet~\cite{deng2009imagenet}, which contains about 9K images for training and 4K images for testing.

\textbf{Input encoding.} To apply the Transformer meta-learner to the task of image regression, we will need to encode the given input of the target image to a set of tokens as the Transformer's data tokens. To achieve this, we follow the practice in Vision Transformer (ViT)~\cite{dosovitskiy2021an} that split the input image into patches. Specifically, the input image is represented by a set of patches of shape $P\times P$, which are converted to flattened vectors $\{p_i\}_{i=0}^{n_p-1}$ with dimension $P\times P\times 3$ for RGB images. For each patch $p_i$, it is assigned with a learnable positional embedding $e_i$. The $i$-th data token is obtained by $\textrm{FC}(p_i + e_i)$ with a FC layer.

\textbf{Implementation details.} On the Imagenette dataset, we apply RandomCrop data augmentation for training our Transformer meta-learner. For both CelebA and Imagenette datasets, the resolution of target images is $178\times 178$ which follows prior practice. We apply a zero-padding of 1 to get the input resolution $180\times 180$, and split the image with patch size $P=9$. For INR, we follow the same 5-layer MLP structure as in prior work~\cite{tancik2020meta}, which has the hidden dimension of 256. The number of groups in weight grouping is 64 by default for a good balance in performance and efficiency. The Transformer follows a similar structure as ViT-Base~\cite{dosovitskiy2021an}, but we reduce the number of layers by half to 6 layers for efficiency. The networks are trained end-to-end with Adam~\cite{kingma2014adam} with a learning rate $1\cdot 10^{-4}$ and the learning rate decays once by 10 when the loss plateaus.

\textbf{Qualitative results.} We first show qualitative results in Figure~\ref{fig:imgrec}. We observe that the Transformer meta-learners are surprisingly effective for building INRs of images in high precision, that can even recover the fine details in complex real-world images. For example, the left example from CelebA shows that our inferred INR $f_\theta$ can successfully reconstruct various details in a face image, including the teeth, lighting effect, and even the background patterns which is not a part of faces. From the right figure of Imagenette dataset, we observe that our inferred INR can recover the digital texts on the object with high fidelity. While it is observed in prior work~\cite{sitzmann2019siren} that learning a latent space of vectors and decoding INRs by hypernetwork can not recover the details in a face image, we show that an INR with precise information can be directly built by a Transformer without any gradient computation.

\begin{table}[t]
    \centering
    \begin{tabular}{ccc}
        \toprule
        & ~~~~~~~~~~CelebA~~~~~~~~~~ & ~~~~Imagenette~~~~ \\
        \midrule
        Learned Init~\cite{tancik2020meta} & 30.37 & 27.07 \\
        Ours & \textbf{31.96} & \textbf{29.01} \\
        \bottomrule
    \end{tabular}
    \caption{\textbf{Quantitative results of image regression (PSNR).} Learned Init is a gradient-based meta-learning algorithm that adapts to an image with a few gradient steps.}
    \label{tab:imgrec}
\end{table}

\begin{table}[t]
    \centering
    \begin{tabular}{cccc}
        $G=1$ & $G=4$ & $G=16$ & $G=64$ \\
        ~{\includegraphics[width=.17\linewidth]{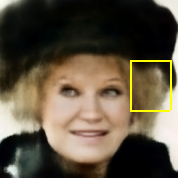}}~ & ~{\includegraphics[width=.17\linewidth]{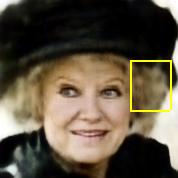}}~ &
        ~{\includegraphics[width=.17\linewidth]{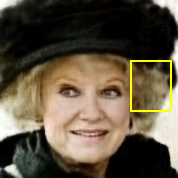}}~ & ~{\includegraphics[width=.17\linewidth]{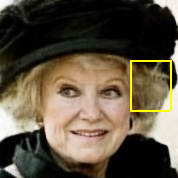}}~
    \end{tabular}
    \begin{tabular}{c|cccc}
        \toprule
        Num of groups ($G$) & 1 & 4 & 16 & 64 \\
        \midrule
        PSNR & ~~~~25.63~~~~ & ~~~~27.89~~~~ & ~~~~29.93~~~~ & ~~~~\textbf{31.96}~~~~ \\
        \bottomrule
    \end{tabular}
    \caption{\textbf{Ablations on the number of weight groups.} The PSNR is evaluated on CelebA dataset. Having more groups ($G$) in weight grouping will make the output INR more flexible and help for representing the details (the yellow box in the shown example).}
    \label{tab:abl_groups}
    \vspace{-1em}
\end{table}

\textbf{Quantitative results.} In Table~\ref{tab:imgrec}, we show quantitative results of our Transformer meta-learner and compare our performance with the gradient-based meta-learning algorithm Learned Init proposed in prior work. Learned Init meta-learns an initialization that can be quickly adapted to target images within a few gradient steps. On both real-world image datasets, we observe that our method achieves the PSNR at around 30 for image regression, and our method without any gradient computation outperforms prior gradient-based meta-learning. The gradient steps involve the repeated process of forward and backward passing through the whole INR sequentially, while ours can directly build the INR by forwarding the information into a shallow Transformer. In summary, our method provides a precise yet efficient hypernetwork-based way of converting image pixels to a global neural function as their underlying INR.

\textbf{Ablations on the number of weight groups.} To justify that the Transformer meta-learner learns about building a complex INR, we show by experiments that the number of groups in weight grouping is not redundant. The qualitative and quantitative results are both shown in Table~\ref{tab:abl_groups}. We observe that by increasing the number of groups $G$ from 1 to 64, the recovered details for the target image are significantly improved in vision, and the PSNR is consistently improving by large margins. The results demonstrate the effectiveness of the weight grouping strategy, and it indicates that the Transformer meta-learner can learn about the complex relations between different weights in the INR so that it can effectively build a large set of weights in a structured way to achieve high precision.

\subsection{View Synthesis}

View synthesis aims at generating a novel view of a 3D object with several given input views. Neural radiance field (NeRF)~\cite{mildenhall2020nerf} has been recently proposed to tackle this task by representing the object as an INR that maps from a 3D coordinate and a viewing direction to the density and RGB value. With the volumetric rendering, the generated views of NeRF are differentiable to the INR weights. View synthesis can be then achieved by first fitting INR for the given input views, then rendering the INR from novel views. The goal of a meta-learner is to infer the INR from given input views efficiently, and improve its generalization so that view synthesis can be achieved with fewer input views.

\textbf{Setup.} We perform view synthesis on objects from ShapeNet~\cite{chang2015shapenet} dataset. We follow prior work~\cite{tancik2020meta} which considers 3 categories: chairs, cars, and lamps. For each category, the objects are split into two sets for training and test, where for each object 25 views (with known camera pose) are provided for training. During testing, a random input view is sampled for evaluating the performance of novel view synthesis.

\textbf{Input encoding.} For each input view image, given the known camera pose, we first compute the ray emitted from every pixel for rendering. The emitted ray at each pixel can be represented as a 3D starting point and a 3D direction vector (normalized as a unit vector). With the original image which has RGB channels, we concatenate all the information at every pixel, which gets an extended image with 9 channels. The extended image contains all the information about an input view, therefore, it can be then split into patches and mapped to data tokens in the Transformer meta-learner. This representation naturally generalizes to multiple input views. Since the information of a single view is represented by a set of patches, when multiple input views are available, their data tokens can be simply merged as a set for representing all the observation information for passing into the Transformer.

\textbf{Adaptive sampling.} To improve the training stability, we propose an adaptive sampling strategy for the first training epoch. Specifically, when we sample the pixel locations for computing the reconstruction loss, we ensure that half of them are sampled from the foreground of the image. This is implemented by selecting the non-white pixels as the background in ShapeNet image is white. We found that the training process is stable after having this simple sampling strategy.

\textbf{Implementation details.} In ShapeNet, input views are provided in resolution $128\times 128$. We split input views with patch size $8$ for the Transformer input. We use NeRF as the INR representation, it follows the architecture in \cite{tancik2020meta} which consists of 6 layers with the hidden dimension of 256 and does not use ``coarse'' and ``fine'' models for simplicity. The architecture of the Transformer and the optimizer are the same as the experiments for image regression.

\begin{figure}[t]
    \begin{subtable}{.3\linewidth}
        \centering
        \begin{tabular}{@{}c|c@{}c@{}c@{}}
            \multicolumn{1}{c}{Input} & GT & w/o T. & w/ T. \\
            \includegraphics[width=.21\linewidth]{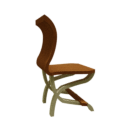} & \includegraphics[width=.21\linewidth]{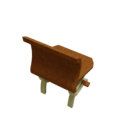} & \includegraphics[width=.21\linewidth]{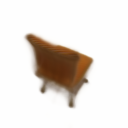} & \includegraphics[width=.21\linewidth]{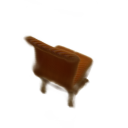} \\
            \includegraphics[width=.21\linewidth]{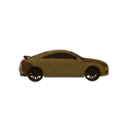} & \includegraphics[width=.21\linewidth]{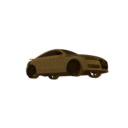} & \includegraphics[width=.21\linewidth]{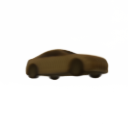} & \includegraphics[width=.21\linewidth]{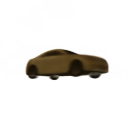} \\
            \includegraphics[width=.21\linewidth]{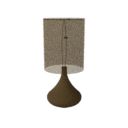} & \includegraphics[width=.21\linewidth]{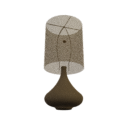} & \includegraphics[width=.21\linewidth]{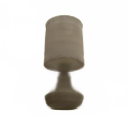} & \includegraphics[width=.21\linewidth]{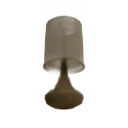}
        \end{tabular}
    \end{subtable}
    \begin{subtable}{.3\linewidth}
        \centering
        \begin{tabular}{@{}c|c@{}c@{}c@{}}
            \multicolumn{1}{c}{Input} & GT & w/o T. & w/ T. \\
            \includegraphics[width=.21\linewidth]{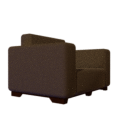} & \includegraphics[width=.21\linewidth]{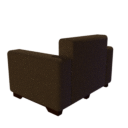} & \includegraphics[width=.21\linewidth]{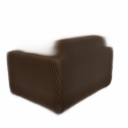} & \includegraphics[width=.21\linewidth]{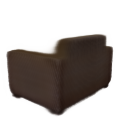} \\
            \includegraphics[width=.21\linewidth]{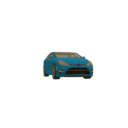} & \includegraphics[width=.21\linewidth]{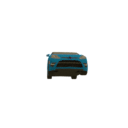} & \includegraphics[width=.21\linewidth]{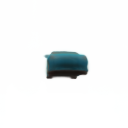} & \includegraphics[width=.21\linewidth]{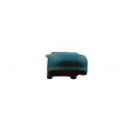} \\
            \includegraphics[width=.21\linewidth]{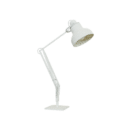} & \includegraphics[width=.21\linewidth]{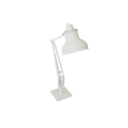} & \includegraphics[width=.21\linewidth]{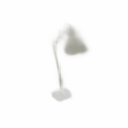} & \includegraphics[width=.21\linewidth]{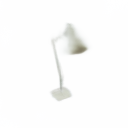}
        \end{tabular}
    \end{subtable}
    \begin{subtable}{.375\linewidth}
        \centering
        \begin{tabular}{@{}c@{}c|c@{}c@{}c@{}}
            \multicolumn{2}{c}{Input} & GT & w/o T. & w/ T. \\
             \includegraphics[width=.168\linewidth]{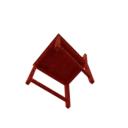} & \includegraphics[width=.168\linewidth]{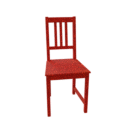} & \includegraphics[width=.168\linewidth]{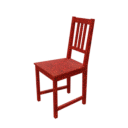} & \includegraphics[width=.168\linewidth]{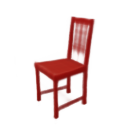} & \includegraphics[width=.168\linewidth]{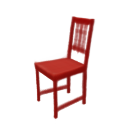} \\
             \includegraphics[width=.168\linewidth]{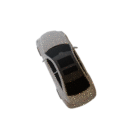} & \includegraphics[width=.168\linewidth]{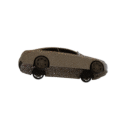} & \includegraphics[width=.168\linewidth]{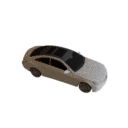} & \includegraphics[width=.168\linewidth]{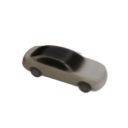} & \includegraphics[width=.168\linewidth]{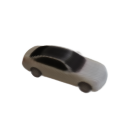} \\ \includegraphics[width=.168\linewidth]{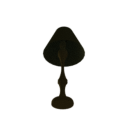} & \includegraphics[width=.168\linewidth]{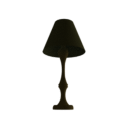} & \includegraphics[width=.168\linewidth]{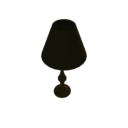} & \includegraphics[width=.168\linewidth]{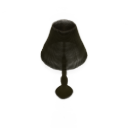} & \includegraphics[width=.168\linewidth]{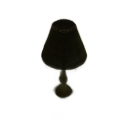}
        \end{tabular}
    \end{subtable}
    \caption{\textbf{View synthesis with Transformer meta-learner on ShapeNet.} The rows show for chairs, cars, and lamps categories. ``w/o T.'' denotes the results of using the Transformer to infer the INR weights without test-time optimization. ``w/ T.'' performs a few test-time optimization steps on the generated INR for the sparse input views, which further helps to reconstruct the fine details in the input views. The corresponding quantitative results are shown in Table~\ref{tab:viewsyn_tto}.}
    \label{fig:viewsyn_tto_visual}
\end{figure}

\begin{table}[t]
    \centering
    \begin{tabular}{cccc}
        \toprule
        & ~~~~Chairs~~~~ & ~~~~Cars~~~~ & ~~~~Lamps~~~~ \\
        \midrule
        NeRF~\cite{mildenhall2020nerf} (Standard~\cite{tancik2020meta}) & 12.49 & 11.45 & 15.47 \\
        Matched~\cite{tancik2020meta} & 16.40 & 22.39 & 20.79 \\
        Shuffled~\cite{tancik2020meta} & 10.76 & 11.30 & 13.88 \\
        \midrule
        Learned Init~\cite{tancik2020meta} & 18.85 & 22.80 & 22.35 \\
        Ours & \textbf{19.66} & \textbf{23.78} & \textbf{22.76} \\
        \bottomrule
    \end{tabular}
    \caption{\textbf{Comparison of building INR for single image view synthesis (PSNR).} The compared methods are baselines and the gradient-based meta-learning algorithm in prior work. Ours does not perform test-time optimization.}
    \vspace{-1em}
    \label{tab:cmp_viewsyn}
\end{table}

\begin{table}[t]
    \centering
    \begin{tabular}{ccccc}
        \toprule
        & \multicolumn{2}{c}{1-view} & \multicolumn{2}{c}{2-view} \\
        & ~~~~~~~w/o T.~~~~~~~ & ~~~~~~~w/ T.~~~~~~~ & ~~~~w/o T.~~~~ & ~~~~w/ T.~~~~ \\
        \midrule
        ~~Chairs~~ & 19.66 & \textbf{20.56} & 21.10 & \textbf{23.59} \\
        Cars & 23.78 & \textbf{24.73} & 25.45 & \textbf{27.13} \\
        Lamps & 22.76 & \textbf{24.71} & 23.11 & \textbf{27.01} \\
        \bottomrule
    \end{tabular}
    \caption{\textbf{Effect of test-time optimization and more input views for view synthesis (PSNR).} We observe that our method for view synthesis can take benefits from test-time optimization and more views.}
    \vspace{-1em}
    \label{tab:viewsyn_tto}
\end{table}

\textbf{Results.} We first compare our method to the prior gradient-based meta-learning algorithm of building INR for single image view synthesis, the results are shown in Table~\ref{tab:cmp_viewsyn}. Standard, Matched, and Shuffled are the baselines trained from different initializations from the prior work~\cite{tancik2020meta}. Specifically, Standard denotes a random initialization, Matched is the initialization learned from scratch which matches the output of the meta-learned initialization, Shuffled is permuting the weights in the meta-learned initialization. We observe that our method outperforms the baselines and the gradient-based meta-learning algorithm for inferring the weights in an INR.

Our method can also naturally take benefits from test-time optimization and more input views. The qualitative and quantitative results are shown in Figure~\ref{fig:viewsyn_tto_visual} and Table~\ref{tab:viewsyn_tto}. We observe that the Transformer meta-learner can effectively build the INR of a 3D object with sparse input views. Since our method builds the whole INR, we can perform further test-time optimization on the INR with given input views just as the original training in NeRF. For efficiency, our test-time optimization only contains 100 gradient steps, it further helps for constructing the fine details in input views. Since the Transformer takes a set as the input, it can gather the information from multiple input views, and we observe the performance can be improved in the setting with more input views.

\section{Does the INR Exploit Data Structures?}
\label{sec:does_inr}

A key potential advantage of INRs is that their representation capacity does not depend on grid resolution but instead on the capacity of the neural network, which allows it to exploit the underlying structures in data and reduce the representation redundancy. To explore whether the structure is modeled in INRs, we visualize the attention weights at the last layer from the weight tokens to the data tokens. Intuitively, since each data token corresponds to a patch in the original image, the attention weights may represent that, for weight columns in different layers, which part of the original image they mostly depend on.

\begin{figure}[t]
    \centering
    \begin{tabular}{c|cccc}
        Image & Layer 1 & Layer 1 & Layer 2 & Layer 3 \\
        \includegraphics[width=.16\linewidth]{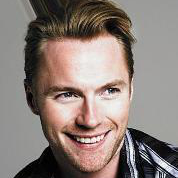} & \includegraphics[width=.16\linewidth]{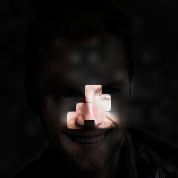} & \includegraphics[width=.16\linewidth]{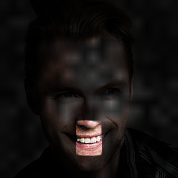} & \includegraphics[width=.16\linewidth]{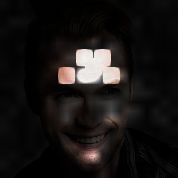} & \includegraphics[width=.16\linewidth]{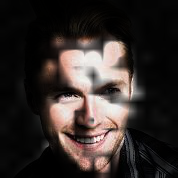} \\
        \includegraphics[width=.16\linewidth]{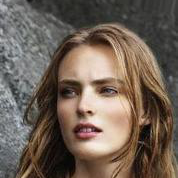} & \includegraphics[width=.16\linewidth]{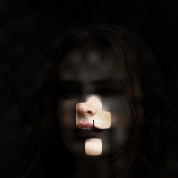} & \includegraphics[width=.16\linewidth]{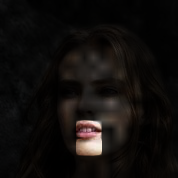} & \includegraphics[width=.16\linewidth]{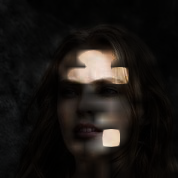} & \includegraphics[width=.16\linewidth]{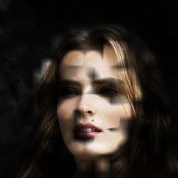}
    \end{tabular}
    \caption{\textbf{Attention masks from weight tokens to data tokens.} Representative examples are selected from tokens for different INR layers. The attention map shows where the corresponding INR weight is attending to.}
    \vspace{-1em}
    \label{fig:vis_attn}
\end{figure}

We reshape the attention weights to the 2D grid of patches and bilinearly up-sample the grid to a mask with the same resolution as the input image. We mask the input image so that parts with higher attention will be shown, the visualization results on CelebA dataset are shown in Figure~\ref{fig:vis_attn}. We observe that there exist weight columns in different layers that attend to structured parts. For example, there are weights roughly attending to the nose and mouth in layer 1, the forehead in layer 2, and the whole face in layer 3. Our observations suggest that the generated INRs may potentially capture the structure of data, which is different from the grid-based discrete representation, where every entry independently represents a pixel and the data structure is not well exploited.

\section{Conclusion}

In this work, we proposed a simple and general formulation that uses Transformers as meta-learners for building neural functions of INRs, which opens up a promising direction with new possibilities. While most of the prior works of hypernetworks for INRs are based on single-vector modulation and high precision reconstruction as a global INR function was mostly achieved by gradient-based meta-learning, our proposed Transformer hypernetwork can efficiently build an INR in one forward pass without any gradient steps, and we observed it can outperform the previous gradient-based meta-learning algorithms for building INRs in the tasks of image regression and view synthesis. While gradient information is not necessary for our model, our method simply builds the weights in a standard INR, therefore it is also flexible to be further combined with any INRs that involve test-time optimization.

Our method draws connections between the Transformer hypernetworks and the gradient-based meta-learning algorithms, and our further analysis sheds light on the generated INRs. We observed that the INR which represents data as a global function may potentially capture the underlying structures without any explicit supervision. Understanding and utilizing these encoded structures could be a promising direction for future works.

~

\textbf{Acknowledgement.} This work was supported, in part, by grants from DARPA LwLL, NSF CCF-2112665 (TILOS), NSF 1730158 CI-New: Cognitive Hardware and Software Ecosystem Community Infrastructure (CHASE-CI), NSF ACI-1541349 CC*DNI Pacific Research Platform, and gifts from Meta, Google, Qualcomm and Picsart.

%
%
\bibliographystyle{splncs04}
\bibliography{main}
\end{document}